\documentclass[11pt]{article}
\usepackage{acl2012}
\usepackage{times}
\usepackage{url}
\usepackage{latexsym}
\usepackage{amssymb}
\usepackage{graphicx}

\title{Multilingual Word Embeddings using Multigraphs}

\author{
  Radu Soricut \\
  Google Inc. \\
  {\tt rsoricut@google.com} \\\And
  Nan Ding \\
  Google Inc. \\
  {\tt dingnan@google.com} \\}

\date{}

\begin{document}
\maketitle
\begin{abstract}
We present a family of
neural-network--inspired models for computing continuous word representations,
specifically designed to exploit both monolingual and multilingual text.
This framework allows us to perform unsupervised training of embeddings that exhibit
higher accuracy on syntactic and semantic compositionality, as well as
multilingual semantic similarity, compared to previous models trained in an unsupervised fashion.
We also show that such multilingual embeddings, optimized for semantic similarity,
can improve the performance of statistical machine translation with respect to how
it handles words not present in the parallel data.
\end{abstract}

\section{Introduction}
Word embeddings are representations that use vectors to represent word surface forms.
They are known to be useful in improving the performance of many NLP tasks,
from sequence labeling~\cite{huang-et-al:2014,turian-et-al:2010},
to language modeling~\cite{bengio-et-al:2003,mnih-et-al:2009},
to parsing~\cite{finkel-et-al:2008,bansal-et-al:2014} and
morphological analysis~\cite{soricut-och:2015,cotterell-etal:2016}.
Cross-lingual tasks (parsing, retrieval, translation) have been particularly attractive
as applications for word embeddings, see~\cite{upadhyay-etal:2016} for a good overview.
Continuous word embeddings use real-valued vectors, and are typically induced either via neural networks~\cite{bengio-et-al:2003,mnih-et-al:2009,socher-et-al:2011a},
or neural-network--inspired models~\cite{mikolov-et-al:2013a,levy-goldberg:2014,pennington-et-al:2014,yu-dredze:2014,ling-etal:2015},
designed to learn word embeddings in an unsupervised manner.

The embeddings that result from training such models exhibit certain desirable properties.
One such property is {\it syntactic compositionality}, i.e.,
the extent to which morpho-syntactic properties (such as pluralization, past-tense, etc.) can be represented
as vectors in the same embedding space, using simple vector arithmetic to encode the application of this property,
e.g., $\mathrm{cars - car + fireman = firemen}$.
Another property is {\it semantic compositionality}, i.e.,
the extent to which semantic properties (such as ``maleness'', ``royalty'', etc.) can be represented
as vectors in the same embedding space, using simple vector arithmetic to encode the application of such a property,
e.g., $\mathrm{king - man + woman = queen}$~\cite{mikolov-et-al:2013b}.
Another desired property, {\it semantic similarity}, means that words with similar meaning are represented as vectors
that are close (cosine-wise) in the embedding space.
Together with semantic compositionality, it allows for manipulating semantic concepts as points in such an embedding space.
The extension of the semantic similarity property to spaces that embed words from more than one language
results in {\it multilingual semantic similarity}, i.e., the extent to which words that are translations of each other
are embedded to vectors that are close in the embedding space~\cite{mikolov-et-al:2013c,gouws-et-al:2015,luong-et-al:2015}.

\begin{figure*}[!t]
\centerline{\includegraphics[width=0.70\textwidth]{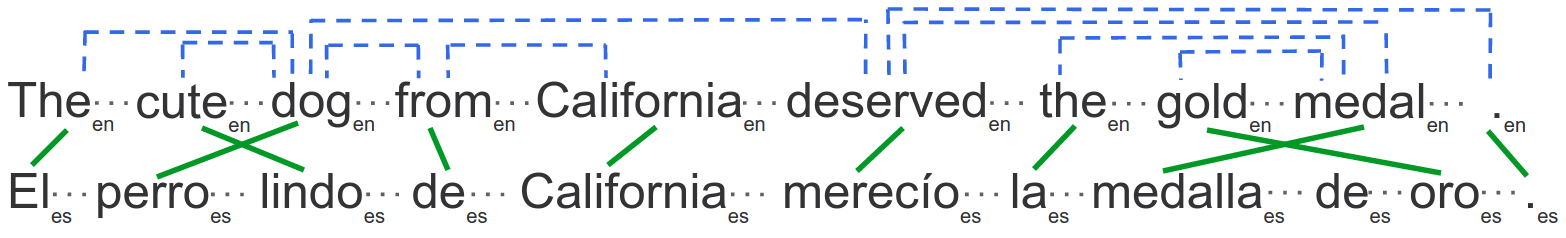}}
\caption{Multigraph with three types of edges}
\label{fig:ex2}
\end{figure*}

In this paper, we present a unified framework for designing neural-network--inspired embedding models using multigraphs.
The new framework introduces the flexibility to exploit
both monolingual and bilingual text, without and with annotations (e.g., syntactic dependencies, word alignments).
Furthermore, it unifies existing models such as the SkipGram model of Mikolov et al.~\shortcite{mikolov-et-al:2013a},
the Dependency embedding model of Levy and Goldberg~\shortcite{levy-goldberg:2014},
and the BiSkip model of Luong et al.~\shortcite{luong-et-al:2015}
as particular instances of this framework.
Our empirical evaluations show that this framework allows us to build models that yield embeddings with
significantly higher accuracy on syntactic compositionality, semantic compositionality,
and multilingual semantic similarity.
We find that both syntactic and semantic compositionality
accuracy of certain embedding spaces improves for a given language due to the presence
of words from other language(s), embedded by the same model in a common embedding space.

We also show that multilingual embedding spaces optimized for semantic similarity
improve the end-to-end performance of a statistical MT system for English-Spanish, English-French, and English-Czech translations.
In order to isolate the impact of this method on end-to-end translation performance,
we perform these experiments on non-neural, phrase-based MT systems with standard dense features~\cite{koehn-etal:03},
rather than the more recently-proposed neural MT systems~\cite{gnmt:16,baidu-mt:16}.
The point of this evaluation is not to improve state-of-the-art in MT;
rather, it allows for a simple way to evaluate how good the translations induced by our multilingual embeddings are.
To this end, we measure the impact of using our approach on handling words that are not present in
the parallel data on which our MT systems are trained.
Potential translations for such words are automatically identified from our multilingual embeddings,
and are subsequently used at decode time to produce improved translations for test sentences
containing such words.
Although the {\em method} used to measure the quality of the induced translations (based on enhancing phrase-tables)
cannot be directly used for neural MT systems, the {\em mapping} of surface strings to their embeddings
in a common multilingual embedding space {\em can} be used directly in fully-neural MT systems,
as an orthogonal method to previously proposed ones, e.g. sub-word unit translations~\cite{sennrich-etal:16,gnmt:16}.

\section{Multigraph-based Embedding Models}
We describe here multigraph-based embedding models, using a formalism that extends on the SkipGram model~\cite{mikolov-et-al:2013a}.
This extension is achieved by replacing the definition of context in SkipGram with the formal definition of multigraph neighborhoods.
We first revisit the SkipGram model, then formally define the notion of multigraph and multigraph-based embedding models.

\subsection{The SkipGram Embedding Models}
The SkipGram neural embedding model of Mikolov et al.~\shortcite{mikolov-et-al:2013a} uses
a vocabulary $V$ and an $n$-dimensional real-valued space $\mathbb{R}^n$.
For each word $w\in V$, a vector $v_w\in \mathbb{R}^n$ is associated as the embedding of the word $w$, learned as follows.

We define $T^w_k$ to be the set of context words of $w$, which is a window of $k$ words to the left and right of $w$. $\overline{T^w_k}$ is the complement for $T^w_k$.
For example, in the English text (1st line) of Figure~\ref{fig:ex2}, $T^{\mathrm{dog_{en}}}_2 = \{\mathrm{The_{en}, cute_{en}, from_{en}, California_{en}} \}$.

The training objective is that for all words $w$ in the training data, the dot product $v_w\cdot v_c$ is
maximized for $v_c \in T_k^w$ while minimized for $v_c \in \overline{T_k^w}$. Formally, this is defined as:
\footnotesize
\begin{eqnarray*}
\arg\max_{\mathbf{v}} \sum_w\sum_{(w,c)\in T^w_k}\hspace*{-4mm}\log\sigma(v_w\cdot v_c) + \hspace*{-4mm}\sum_{(w,c)\in \overline{T^w_k}}\hspace*{-4mm}\log\sigma(-v_w\cdot v_c),
\end{eqnarray*}
\normalsize
\noindent
where $\sigma(x) = \frac{1}{1 + e^x}$. Due to its large size, $\overline{T^w_k}$ is approximated via sampling:
for each $(w,c)\in T^w_k$, we draw $n$ samples $(w,c_i)\not\in T^w_k$ according to the
distribution $U(c_i)^{3/4}/Z$ ($U$ is the unigram distribution, $Z$ the normalization constant), where $n$ is a model hyper-parameter.

\subsection{Multigraph-based Embedding Models}
The SkipGram embedding models only make use of the context windows. However, there is much more information in both monolingual and bilingual text, e.g. syntactic dependencies and word alignments. For example, in Figure~\ref{fig:ex2}, besides the dotted edges connecting neighbors in the text surface (denoted as $\mathrm{T}$ edges), the dashed edges connect words that are in a syntactic dependency relation (denoted as $\mathrm{D}$ edges), and solid edges connect nodes that represent translation-aligned words (denoted as $\mathrm{A}$ edges).

In order to make use of all three types of information, we first define a set of labels $L = \{\mathrm{T, A, D}\}$,
and a distance function $d: L\rightarrow \mathbb{N}$ by assigning a distance to each label.
We use notation $\mathrm{T_iA_jD_k}$ to represent $d(\mathrm{T})=i$, $d(\mathrm{A})=j$, $d(\mathrm{D})=k$.
If any of the labels are mapped to distance 0, we omit it from the notation;
for instance, $\mathrm{T_5A_1}$ means $d(\mathrm{T})=5$, $d(\mathrm{A})=1$, $d(\mathrm{D})=0$.

Next, we show how to incorporate those edges with different labels and distance functions into the same embedding model using multigraphs.

\subsubsection{Multigraphs}
A multigraph is a graph which is allowed to have multiple edges between the same nodes.
We use here the particular notion of an undirected multigraph with
identifiable edges (i.e., any edge has an identity that helps it being distinguished from
a different edge that may share the same end nodes).
Formally, an undirected multigraph with
identifiable edges is a 5-tuple $G:=(V, E, L, F, I)$, where
$V$ is a set of vertices or nodes; $E$ is a set of edges; $L$ is a set of labels;
$F: E\rightarrow V\times V$ assigns end nodes to edges; and
$I: E\rightarrow L$ assigns identity labels to edges.

We use notation $G^w_d$ to represent the neighborhood of node $w$ in $G$ as given by a distance function
$d$. $G^w_d$ is defined as:
\[
G^w_d = \{ (w,c) | \exists w\stackrel{e_1 \to \ldots \to e_k}{\longrightarrow} c ; k \leq \min_{l\in \{ L(e_i) \}_{i=1}^k}\hspace*{-4mm} d(l) \}
\]
\noindent
where $e_1 \to \ldots \to e_k$ denotes the edges of a length-$k$ path from $w$ to $c$.

Using the example from Figure~\ref{fig:ex2}, we have $\mathrm{dog_{en}} \in G^{\mathrm{medal_{en}}}_{T_1D_2}$, because there is a length-2 path with two $D$-type edges $(\mathrm{medal_{en}},\mathrm{deserved_{en}}), (\mathrm{deserved_{en}},\mathrm{dog_{en}})$, with $\min d(l) = d(A) = 2$ and $k = 2$. On the other hand, $\mathrm{California_{en}} \not\in G^{\mathrm{medal_{en}}}_{T_1D_2}$, because the path contains an $D$-type edge $(\mathrm{medal_{en}},\mathrm{deserved_{en}})$ and a $T$-type edge $(\mathrm{deserved_{en}},\mathrm{California_{en}})$, hence $\min d(l) = d(T) = 1$ while $1 < k$.

\subsubsection{Objective Function}
Training the embedding variables $\mathbf{v}$ is based on the neighborhoods $G_d^w$ for all words $w$ in the multigraph. The general form of the objective function can take any arbitrary function defined on $G_d^w$ and $\overline{G^w_d}$, where $\overline{G^w_d}$ is the complement of $G^w_d$. We use here a similar objective function to the one used by the SkipGram model:
\footnotesize
\begin{eqnarray}
\arg\max_{\mathbf{v}} \sum_w\sum_{(w,c)\in G^w_d}\hspace*{-4mm}\log\sigma(v_w\cdot v_c) +\hspace*{-4mm} \sum_{(w,c)\in \overline{G^w_d}}\hspace*{-4mm}\log\sigma(-v_w\cdot v_c)\label{eq:obj}
\end{eqnarray}
\normalsize
\noindent
where $\sigma(x) = \frac{1}{1 + e^x}$. Due to its large size, $\overline{G^w_d}$ is approximated similarly with $\overline{T_k^w}$:
for each $(w,c)\in G^w_d$, we draw $n$ samples $(w,c_i)\not\in G^w_d$ according to the
unigram~\footnote{The unigram vocabulary over multilingual text is considered to be the union over all language-tagged word types.}
distribution $U(c_i)^{3/4}/Z$, where $n$ is a model hyper-parameter.

\subsubsection{Discussion}
Given sentence-level parallel data with word alignments,
Equation~\ref{eq:obj} can be used to learn parameters for a model such as $\mathrm{T_1A_1}$.
Under this model, the graph neighborhood for $\mathrm{dog_{en}}$ in Figure~\ref{fig:ex2} is
$G^{\mathrm{dog_{en}}}_{\mathrm{T_1A_1}} =$
\small$\{ \mathrm{cute_{en}, from_{en}, perro_{es}} \}$\normalsize.
The neighborhood for $\mathrm{perro_{es}}$ is $G^{\mathrm{perro_{es}}}_{\mathrm{T_1A_1}} =$
\small$\{ \mathrm{El_{es}, lindo_{es}, dog_{en}} \}$\normalsize.

If syntactic dependency relations are available (for one or both languages in the parallel data),
we can also learn parameters for model $\mathrm{T_1A_1D_1}$.
For the multigraph in Figure~\ref{fig:ex2}, the graph neighborhood for $\mathrm{dog_{en}}$ is
$G^{\mathrm{dog_{en}}}_{\mathrm{T_1A_1D_1}} =$
\small$\{ \mathrm{The_{en}, cute_{en}, from_{en}, deserved_{en}, perro_{es}} \}$\normalsize.

We can also create multigraph models for which the set of labels $L$ collapses some of the labels.
For instance, collapsing labels T and A leads to a multigraph model such as
$\mathrm{(TA)_3}$, under which both text-based edges and alignment-based edges are traversed
without differentiation (up to distance 3).
In this case, the graph neighborhood for $\mathrm{dog_{en}}$ is
$G^{\mathrm{dog_{en}}}_{\mathrm{(TA)_3}} =$
\small
$\{ \mathrm{The_{en}, cute_{en}, from_{en}, California_{en}, deserved_{en}, El_{es},}$
$\mathrm{perro_{es}, lindo_{es}, de_{es}, California_{es}} \}$.
\normalsize
For $\mathrm{perro_{es}}$, it is
$G^{\mathrm{perro_{es}}}_{\mathrm{(TA)_3}} = $
\small
$\{ \mathrm{El_{es}, lindo_{es}, de_{es}, California_{es},}$
$\mathrm{The_{en}, cute_{en}, dog_{en}, from_{en}, California_{en}} \}.$
\normalsize

Note the ~90\% overlap between $G^{\mathrm{dog_{en}}}_{\mathrm{(TA)_3}}$ and $G^{\mathrm{perro_{es}}}_{\mathrm{(TA)_3}}$.
Since the objective function from Equation~\ref{eq:obj} imposes that words that appear in similar contexts
have similar embeddings, it follows that, under a model like $\mathrm{(TA)_3}$,
the embeddings for $\mathrm{dog_{en}}$ and $\mathrm{perro_{es}}$ should be similar.
This leads to embeddings for words in multiple languages that keeps translations close to each other.
On the other hand, models like $\mathrm{T_1A_1}$ and $\mathrm{T_1A_1D_1}$ do not have this property,
but nevertheless give rise to embeddings for words in multiple languages with properties
that we analyze and quantify in Section~\ref{sec:evaluation}.

\section{Related Work}
Distributed word representations have been used recently for tackling various
tasks such as language modeling~\cite{mnih-hinton:2007,mikolov-zweig:2012},
paraphrase detection~\cite{socher-et-al:2011a},
sentiment analysis~\cite{socher-et-al:2011c},
syntactic parsing~\cite{collobert:2011,faruqui-dyer:2014,guo-etal:2015,huang-etal:2015},
and a multitude of cross-lingual tasks~\cite{upadhyay-etal:2016}.

The introduction of the CBOW and SkipGram embedding models~\cite{mikolov-et-al:2013a} has boosted this research direction.
These models are simple and easy to implement, and can be trained orders of magnitude faster than previous models.
Subsequent research has proposed models that take advantage of additional textual information,
such as syntactic dependencies~\cite{levy-goldberg:2014,bansal-et-al:2014}, global statistics~\cite{pennington-et-al:2014},
or parallel data~\cite{mikolov-et-al:2013c,gouws-et-al:2015,luong-et-al:2015,mogadala-rettinger:2016}.
Prior knowledge can also be incorporated to achieve improved lexical embeddings
by modifying the objective function while allowing for the exploitation of existing resources such as WordNet~\cite{yu-dredze:2014},
or by modifying the model architecture while targeting specific tasks~\cite{ling-etal:2015}.

Our paper describes a mechanism which unifies the way context signals from the training data are exploited.
It exploits the information available in parallel data in an on-line training fashion (bi/multi-lingual training),
compared to the off-line matrix transformation proposal of Mikolov et al.~\shortcite{mikolov-et-al:2013c}.
The BilBOWA model~\cite{gouws-et-al:2015} uses a parallel bag-of-word representation for the parallel data,
while the BiSkip model~\cite{luong-et-al:2015} achieves bilingual training by exploiting word alignments.
Some of these proposals can be formulated as particular instances under our multigraph framework.
For instance, the context window (with window size $k$) $T_k^w$ of the SkipGram model is equivalent to model $G^w_{T_k}$ in the multigraph formulation.
The dependency-based embedding model of Levy and Goldberg~\shortcite{levy-goldberg:2014} is equivalent to $G^w_{D_1}$ when word and context vocabulary are the same~\footnote{
We also ignore the collapse of preposition-based dependencies, which
makes their model be between $\mathrm{D_1}$ and $\mathrm{D_2}$.}.
The BiSkip model~\cite{luong-et-al:2015} with a window of size $k$ is equivalent to model $G^w_{(TA)_k}$ in our multigraph formulation.

In addition to subsuming some of the previously-proposed methods, our approach comes with a mathematical foundation (in terms of multigraphs) for incorporating information from both monolingual and parallel data.
This formulation allows us to understand and justify, from a formal perspective, some of the empirical results obtained by some of these models.
Moreover, our method allows for the exploitation of signals obtained via both unsupervised (e.g. raw text, parallel text with
unsupervised alignments) and supervised learning (e.g., syntactic dependencies), while
building a common embedding over arbitrary many languages, simply by treating the training data as a multigraph over potentially multiple languages, linked together via multiple bilingual alignments.

A related approach for inducing multilingual embeddings is based on neural networks
for automatic translation, either in conjunction with a phrase-table~\cite{devlin-etal:2014,devlin-part2:2015},
or a fully neural approach~\cite{sutskever-etal:2014,cho-etal:2014,gnmt:16,baidu-mt:16}.
These approaches use signals similar to ours when exploiting parallel training data,
but the resulting embeddings are optimized for translation accuracy
(according to the loss-function definition of these models, usually using a
maximum-likelihood objective~\cite{sutskever-etal:2014,cho-etal:2014}
or a reinforcement-learning--inspired objective~\cite{gnmt:16}).
In addition, they do not directly allow for the exploitation of both parallel and monolingual data simultaneously at train time,
or the exploitation of additional sources of linguistic information (such as syntactic dependencies).

Because our approach enables us to exploit both monolingual and parallel data simultaneously,
the resulting distributed representation can be successfully used to learn translations for terms that appear in
the monolingual data only (Section~\ref{sec:ex-eval}).
This represents the neural word-embedding equivalent of a long line of research based on word-surface patterns,
starting with earlier attempts~\cite{rapp:1995}, and continuing with~\cite{koehn-knight:2002,garera-et-al:2009}, and complemented by
approaches based on probabilistic models~\cite{haghighi-et-al:2008}.
Our approach has the advantages of achieving this effect in a completely unsupervised fashion, without exploiting surface patterns, and
benefiting from the smoothness properties associated with continuous word representations.

\begin{table*}[th]
\begin{center}
\begin{tabular}{rrcc}
 Test Name   & Input                & Reference & Size \\ \hline
Word-Rel Syn & car - child + children & cars & 10,675 \\
Word-Rel Sem & policeman - groom + bride & policewoman & 8,869 \\
Mitchell   & $\langle$elderly \& woman, black \& hair$\rangle$ & 1.6 & 324 \\
           & $\langle$amount \& reduce, cost \& cut$\rangle$   & 6.6 & \\
Stanford-C & $\langle$... is a celebrated \underline{jazz} clarinetist ..., & & 2,003 \\
           & ... an English \underline{rock} band ...$\rangle$ & 8.2 & \\
Stanford-R & $\langle$amorphous, inorganic$\rangle$            & 1.9 & 2,034 \\
           & $\langle$belligerence, hostility$\rangle$         & 8.7 & \\
Mikolov Transl. EnEs & strained  & tenso & 1,000 \\
Mikolov Transl. EsEn & inteligente & clever & 1,000 \\ \hline
\end{tabular}
\caption{Descriptions of the standard test sets used in our evaluations.}
\label{tab:testsets}
\end{center}
\end{table*}

\def\ty{\tiny}
\def\ss{\scriptsize}
\def\fn{\footnotesize}
\def\sm{\small}
\def\sp{$\;$}
\begin{table*}[th]
\begin{center}
\begin{tabular}{rcccccc}
      & Word-Rel Syn  & Word-Rel Sem & & Mitchell & Stanford-C & Stanford-R \\ \cline{2-7}
Model &   \multicolumn{2}{c}{Acc@1} & & \multicolumn{2}{c}{Spearman $\rho$} \\ \cline{1-3}\cline{5-7}
(SkipGram) $\mathrm{T_5}$    & 68.4\sp\ss[67.5, 69.2] & 76.7\sp\ss[75.8, 77.6] & & 64.0\sp\ss[58.0, 70.8] & 66.5\sp\ss[63.9, 69.2] & 42.6\sp\ss[39.0, 46.3] \\
(Dependency) $\mathrm{D_1}$  & 67.2\sp\ss[66.3, 68.1] & 46.4\sp\ss[45.4, 47.4] & & 65.3\sp\ss[59.8, 71.6] & 66.7\sp\ss[64.3, 69.5] & 41.1\sp\ss[37.4, 44.8] \\
           $\mathrm{T_5D_1}$ & \bf 72.6\sp\ss[71.8, 73.5] & 76.2\sp\ss[75.3, 77.0] & & 65.3\sp\ss[59.8, 71.6] & 68.3\sp\ss[65.9, 70.9] & 45.2\sp\ss[41.7, 48.8] \\ \hline
\end{tabular}
\caption{Evaluation for models $\mathrm{T_5}$, $\mathrm{D_1}$, $\mathrm{T_5D_1}$,
trained on English Wikipedia (1Bw). $\mathrm{T_5D_1}$ is significantly better on Word-Rel Syn, while all other tests stay at similar levels.
}
\label{tab:eval1}
\end{center}
\end{table*}

\section{Empirical Results}
\label{sec:evaluation}
In this section, we present empirical evaluations aimed at answering several questions:
How do multigraph embeddings, such as $\mathrm{T_1A_1}$, $\mathrm{T_1A_1D_1}$, $\mathrm{(TA)_5}$,
differ from regular SkipGram embeddings (i.e., $\mathrm{T_5})$?
What leverage do we get from using multilingual data?
How can we use multigraph embeddings to accomplish non-trivial end-to-end tasks (such as automatic translation)?
We provide answers for each of these questions next.

\subsection{Data}
\label{sec:data}
As training data, we use publicly available parallel and monolingual data.
For parallel data, we use the Spanish-English, French-English, German-English, Russian-English,
and Czech-English data released as part of the WMT-2015 shared task~\cite{WMT:2015}.
For monolingual data, we use the LDC Gigaword collections for English (5 Bw)~\cite{ldc-english-gigaword}, Spanish (1.4 Bw)~\cite{ldc-spanish-gigaword},
and French (1.1 Bw)~\cite{ldc-french-gigaword}.
For monolingual English, we also use the Wikipedia data (1 Bw)~\cite{Westbury-Wikipedia}.
This setup allows us to directly compare with previous published results that use the same data sets.

As test data, we use several standard, publicly released test sets, see Table~\ref{tab:testsets}.
We use the Semantic-Syntactic Word Relationship dataset introduced
in~\cite{mikolov-et-al:2013b} by reporting results separately on the syntactic and the semantic parts.
The role of the Word-Rel Syn part of this dataset is to quantify the
syntactic compositionality property of an embedding space;
the role of the Word-Rel Sem part is to quantify the
semantic compositionality property of an embedding space.
We also use the Mitchell dataset~\cite{mitchell-lapata:2010}, consisting of pairs of two-word
phrases and a human similarity judgment on the scale of 1-7;
the Stanford English in-context word-similarity (Stanford-C) dataset~\cite{huang-et-al:2012},
consisting of sentence pairs and a human similarity judgment on the scale of 1-10;
the Stanford English Rare-Word (Stanford-R) dataset~\cite{luong-et-al:2013},
consisting of word pairs and human similarity judgments, containing a higher degree of English morphology compared to other
word-similarity datasets;
the Mikolov translation dataset~\cite{mikolov-et-al:2013c}, consisting of English-Spanish
word pairs that are translations of each other (single reference).
All confidence intervals reported are at 95\% confidence level.

\subsection{Intrinsic Evaluations}
\label{sec:in-eval}
As evaluation metrics, we use Acc@1 (the percent of cases for which the 1-highest cosine exactly matches the reference),
Acc@5 (the percent of cases for which one of the 5-highest cosine exactly matches the reference),
and Spearman correlation (how well cosine-based rankings match human-score--based rankings).

\subsubsection*{Two Edge Types are Better than One: $\mathrm{T_5D_1}$}
\noindent
The first multigraph model we evaluate is model $\mathrm{T_5D_1}$.
This model formally subsumes both the SkipGram model
(via $\mathrm{T_5}$ neighborhoods) and the Dependency model (via $\mathrm{D_1}$ neighborhoods).
Table~\ref{tab:eval1} presents the performance of this model across five different testsets,
against models $\mathrm{T_5}$ and $\mathrm{D_1}$.
For training we used English Wikipedia (1B words), and a dependency parser similar with the one
described in~\cite{petrov:2012}.

The results show that model $\mathrm{T_5D_1}$ is capable of combining the strength of both
the $\mathrm{T_5}$ model and the $\mathrm{D_1}$ model, and achieves significantly improved performance on
syntactic compositionality (72.6\% Acc@1 on Word-Rel Syn, compared to 68.3\% for SkipGram).
On all other tests, the performance does not degrade compared to the best achieved by one of the single-edge type models:
76.2\% Acc@1 on Word-Rel Sem, compared to 76.7\% for SkipGram;
68.3 Spearman $\rho$ on Stanford-C, compared to 66.5 for SkipGram, and 65.7 reported by~\cite{huang-et-al:2012} on the same training set;
45.2 Spearman $\rho$ on Stanford-R, compared to 42.6 for SkipGram, and 34.4 reported by~\cite{luong-et-al:2013} on the same training set.

\def\mltwo{EnEs+EnDe}
\def\mlfour{EnEs+EnDe+EnRu+EnCz}

\subsubsection*{Improved syntactic compositionality: $\mathrm{T_1A_1}$}
\noindent
To investigate in more detail the syntactic compositionality property,
we train multigraph models of type $\mathrm{T_5}$ (SkipGram) and $\mathrm{T_1A_1}$ on
various language-pairs, using the WMT parallel datasets, as follows:
1LP=EnEs; 2LPs=\mltwo; 4LPs=\mlfour.
The results in Table~\ref{tab:eval2} show the positive impact on
the syntactic compositionality due to the presence of the other language(s).

\begin{table}[th]
\begin{center}
\begin{tabular}{rcc}
      &    Word-Rel Syn & Word-Rel Sem \\  \cline{2-3}
Model &    \multicolumn{2}{c}{Acc@1} \\ \hline
\sm$\mathrm{T_1}$[1LP]     & 26.9\sp\ss[26.1, 27.7] &  2.5\sp\ss[2.2, 2.9] \\
\sm$\mathrm{T_5}$[1LP]     & 37.8\sp\ss[36.9, 38.6] & 18.3\sp\ss[17.4, 19.1] \\
\sm$\mathrm{T_1A_1}$[1LP]  & \bf 51.6\sp\ss[50.7, 52.5] &  5.9\sp\ss[5.4, 6.4] \\ \hline
\sm$\mathrm{T_5}$[2LPs]    & 48.0\sp\ss[47.1, 48.9] & 33.0\sp\ss[32.0, 34.0] \\
\sm$\mathrm{T_1A_1}$[2LPs] & \bf 57.9\sp\ss[57.0, 58.8] & 11.3\sp\ss[10.6, 11.9] \\ \hline
\sm$\mathrm{T_5}$[4LPs]    & 46.8\sp\ss[45.8, 47.7] & 39.6\sp\ss[38.6, 40.6] \\
\sm$\mathrm{T_1A_1}$[4LPs] & \bf 59.4\sp\ss[58.4, 60.2] & 12.1\sp\ss[11.4, 12.8] \\ \hline
\end{tabular}
\caption{Model $\mathrm{T_5}$ versus $\mathrm{T_1A_1}$ on Word-Rel (Syn/Sem);
multilingual setups with WMT data.}
\label{tab:eval2}
\end{center}
\end{table}

Note that Word-Rel Syn measures properties of embeddings for English words only.
Yet the multilingual aspect of the $\mathrm{T_1A_1}$ embeddings directly impacts the
syntactic compositionality of English embeddings (59.3\% Acc@1 on Word-Rel Syn with the 4LPs combination,
versus 46.8\% Acc@1 when using only the English side of the same training data for $\mathrm{T_5}$).
Also note that this is not due only to simply having more tokens/events when training the models:
for instance, $\mathrm{T_1A_1}$[1LP] is trained on roughly the same number of tokens as $\mathrm{T_5}$[2LPs]
(the former has tokens in English and Spanish, while the latter has only English tokens but on twice more data),
yet the Acc@1 of the former, at 51.6\%, is statistically-significantly better compared to the latter, at 48.0\%.

Another significant finding from Table~\ref{tab:eval2} is that the improved syntactic compositionality
of $\mathrm{T_1A_1}$ embedding spaces comes at the expense of semantic compositionality
(12.1\% Acc@1 on Word-Rel Sem with the 4LPs combination, compared to 39.6\% Acc@1 for $\mathrm{T_5}$).
Poor semantic compositionality appears to be a consequence of $\mathrm{T_1A_1}$ multigraph neighborhoods being too local;
we show in later results how to mitigate this issue.

\subsubsection*{Improved semantic compositionality: $\mathrm{(TA)_5}$}
\noindent
We use the same multilingual training setup to investigate in more detail
the semantic compositionality property as well.
We train multigraph models of type $\mathrm{(TA)_5}$ on the same combination of WMT parallel datasets as above.

\begin{table}[th]
\begin{center}
\begin{tabular}{rcc}
      & Word-Rel Syn & Word-Rel Sem \\ \cline{2-3}
Model & \multicolumn{2}{c}{Acc@1} \\ \hline
\sm$\mathrm{T_5}$[1LP]      & 37.8\sp\ss[36.9, 38.6] & 18.3\sp\ss[17.4, 19.1] \\
\sm$\mathrm{(TA)_5}$[1LP]   & \bf 40.9\sp\ss[40.0, 41.8] & \bf 30.7\sp\ss[29.8, 31.7] \\ \hline
\sm$\mathrm{T_5}$[2LPs]     & 48.0\sp\ss[47.1, 48.9] & 33.0\sp\ss[32.0, 34.0] \\
\sm$\mathrm{(TA)_5}$[2LPs]  & 48.1\sp\ss[47.2, 49.0] & \bf 43.1\sp\ss[42.0, 44.0] \\ \hline
\sm$\mathrm{T_5}$[4LPs]     & 46.8\sp\ss[45.8, 47.7] & 39.6\sp\ss[38.6, 40.6] \\
\sm$\mathrm{(TA)_5}$[4LPs]  & 48.2\sp\ss[47.3, 49.1] & \bf 46.4\sp\ss[45.3, 47.4] \\ \hline
\end{tabular}
\caption{Model $\mathrm{T_5}$ versus $\mathrm{(TA)_5}$ on Word-Rel (Syn/Sem);
multilingual setups with WMT data.}
\label{tab:eval3}
\end{center}
\end{table}

The results in Table~\ref{tab:eval3} show a significant positive impact,
this time on the semantic compositionality of the resulting
embeddings space, due to the presence of the other language(s).
We emphasize that the Word-Rel Sem dataset measures properties of embeddings for English words only,
but the multilingual aspect of the $\mathrm{(TA)_5}$ embeddings directly impacts the
semantic compositionality of the English embeddings
(Word-Rel Sem at 46.4\% Acc@1 under the 4LPs condition, versus 39.6\% for SkipGram).

For this model, the improved semantic compositionality also comes with a statistically-significant
increase in syntactic compositionality
(Word-Rel Syn 48.2\% Acc@1 under the 4LPs condition, versus 46.8\% for SkipGram).
However, this result is significantly below the one obtained by the $\mathrm{T_1A_1}$ model
on the same training data on syntactic compositionality (59.3\% Acc@1 on Word-Rel Syn, see Table~\ref{tab:eval2}).

\begin{table*}[th]
\begin{center}
\begin{tabular}{rcccccc}
      &   Word-Rel Syn & Word-Rel Sem & & Mitchell & Stanford-C & Stanford-R \\ \cline{2-7}
Model &   \multicolumn{2}{c}{Acc@1} & & \multicolumn{3}{c}{Spearman $\rho$} \\ \cline{1-3}\cline{5-7}
\sm(SkipGram) $\mathrm{T_5}$[En+4LPs]            & 67.8\sp\ss[67.0, 68.6] & 74.4\sp\ss[73.4, 75.3] & & 68.4\sp\ss[62.7, 74.3] & 67.4\sp\ss[64.9, 70.1] & 44.8\sp\ss[41.3, 48.4] \\
\sm$\mathrm{T_5D_1}$[En]+$\mathrm{T_1A_1}$[4LPs]  & \bf 72.0\sp\ss[71.2, 72.8] & \bf 78.3\sp\ss[77.4, 79.2] & & 66.4\sp\ss[60.9, 72.9] & 68.6\sp\ss[66.1, 71.3] & 46.6\sp\ss[43.1, 50.1] \\ \hline
\end{tabular}
\caption{Multigraph embeddings in a monolingual plus (multi) bilingual setup.
The monolingual data [En] is Wikipedia (1Bw), the parallel data (4LPs=\mlfour) is WMT-2015 (1Bw on English side).}
\label{tab:eval4}
\end{center}
\end{table*}

\subsubsection*{Improved syntactic+semantic compositionality}
\noindent
It is also possible to train embeddings that achieve superior
performance on both syntactic and semantic compositionality.
To this end, we train a combined model that runs on
$\mathrm{T_5D_1}$ multigraphs on monolingual English data (Wikipedia 1Bw),
plus $\mathrm{T_1A_1}$ multigraphs on parallel data; we denote the resulting model
$\mathrm{T_5D_1}$[En]+$\mathrm{T_1A_1}$[4LPs].

As the evaluation in Table~\ref{tab:eval4} shows,
$\mathrm{T_5D_1}$[En]+$\mathrm{T_1A_1}$[4LPs] registers
both the improved syntactic compositionality of $\mathrm{T_1A_1}$ and
the improved semantic compositionality of $\mathrm{T_5D_1}$.
This model significantly improves on testsets impacted by syntactic compositionality
(72.0\% Acc@1 on Word-Rel Syn versus 68.5\% for SkipGram)
and semantic compositionality
(78.3\% Acc@1 on Word-Rel Sem versus 74.4\% for SkipGram).
We make the conjecture here that better compositionality properties
can improve the end-performance of techniques that explicitly exploit them, such
as morphology induction~\cite{soricut-och:2015}.

\subsubsection*{Improved multilingual embeddings: $\mathrm{(TA)_5}$}
\noindent
We also investigate the extent to which multigraph-based embeddings
encode multilingual semantic similarity.
The intuition for this property is that words in various languages that are translations
of each other should embed to points that are close to each other in the common embedding space.

\begin{table}[th]
\begin{center}
\begin{tabular}{lll}
      &  \multicolumn{2}{c}{Mikolov Translation Test} \\ \cline{2-3}
Model                         & EnEs   & EsEn \\
                              & Acc@1  & Acc@1 \\ \hline
\fn~\cite{mikolov-et-al:2013c}$\!\!\!\!$ & 33.0\sp\ss[--,--]     & 35.0\sp\ss[--,--]   \\
\fn~\cite{gouws-et-al:2015}        & 39.0\sp\ss[--,--]     & 44.0\sp\ss[--,--]   \\
\fn$\mathrm{(TA)_5}$[EnEs]      & \bf 57.3\sp\ss[54.3, 60.3] & \bf 58.2\sp\ss[55.2, 61.0]  \\ \hline
\end{tabular}
\caption{Evaluations of model $\mathrm{(TA)_5}$ on a intrinsic translation accuracy task.
Training data for all models is English-Spanish WMT-2015.}
\label{tab:eval5}
\end{center}
\end{table}

The results in Table~\ref{tab:eval5} show the superiority of $\mathrm{(TA)_5}$ embeddings
compared to embeddings obtained with previously-proposed approaches:
the model of Mikolov et al.~\shortcite{mikolov-et-al:2013c}, and the BilBOWA model of
Gouws et al.~\shortcite{gouws-et-al:2015}.
At 57.3\% Acc@1 on EnEs and 58.2\% Acc@1 on EsEn, model $\mathrm{(TA)_5}$ outperforms by
a large margin the best results reported on these test sets.
The advantage of the $\mathrm{(TA)_5}$ model is that it can fully exploit the information made available by
word-level alignments during on-line training:
under the objective function from Eq.~\ref{eq:obj} using multigraph-based neighborhoods,
words aligned in the training data embed to points that have high cosines in the common embedding space.

\begin{table}[th]
\begin{center}
\begin{tabular}{rcc}
      &  \multicolumn{2}{c}{Mikolov Translation Test} \\
      &  \multicolumn{2}{c}{EnEs}        \\ \cline{2-3}
Model                      & Acc@1                 &  Acc@5      \\ \hline
\sm$\mathrm{(TA)_5}$[EnEs] & \bf 57.3\sp\ss[54.3, 60.3] & 86.3\sp\ss[84.3,88.4] \\
\sm$\mathrm{(TA)_5}$[2LPs] & 51.6\sp\ss[48.7, 54.5] & \bf 90.0\sp\ss[88.1, 91.7] \\
\sm$\mathrm{(TA)_5}$[4LPs] & 49.3\sp\ss[46.2, 52.2] & \bf 90.6\sp\ss[88.9, 92.3] \\ \hline
\end{tabular}
\caption{Intrinsic translation accuracy with model $\mathrm{(TA)_5}$;
multilingual setups with WMT-2015 data.}
\label{tab:eval6}
\end{center}
\end{table}

We also test what happens when we train our models on multiple bilingual training sets.
In addition to training the $\mathrm{(TA)_5}$ model on the English-Spanish parallel data only,
we also train it using two bilingual training sets (2LPs setup, for \mltwo), and
four bilingual training sets (4LPs setup, for \mlfour).
The figures in Table~\ref{tab:eval6} show the results of such a setup.
Note that we tag words from different languages with language tags (see Figure~\ref{fig:ex2}),
and compute cosine ranks (i.e., Acc@1 and Acc@5) considering the embedded points from the desired target language only.
The results show lower translation performance on Acc@1 on the English-Spanish test set:
from 57.3\% Acc@1 for $\mathrm{(TA)_5}$[EnEs] under the 1LP (English-Spanish) condition,
to 51.6\% Acc@1 under the 2LPs condition, to 49.4\% Acc@1 under the 4LPs condition.
Our interpretation is that multiple bilingual events tend to pull the
embedding parameters for English words in different directions, leading to
a decrease in Acc@1 translation accuracy.
However, under Acc@5 this trend is reversed, and accuracy numbers improve when adding more bilingual training data.
This is an indication that, even though the addition of multiple languages inserts some additional noise
to the embeddings (for instance, due to polysemy),
multiple bilingual datasets contribute positively to the creation of a common embedding space that is geared
toward translation invariance.

\def\mtmodel{\small{$\mathrm{T_5}[S]+\mathrm{(TA)_5}[S$-$T]+\mathrm{T_5}[T]$}\normalsize}
\def\mtmodelenes{\sm$\mathrm{T_5}$[En]$+\mathrm{(TA)_5}$[EnEs]$+\mathrm{T_5}$[Es]\normalsize}
\def\mtmodelenfr{\sm$\mathrm{T_5}$[En]$+\mathrm{(TA)_5}$[EnFr]$+\mathrm{T_5}$[Fr]\normalsize}
\def\mtmodelencz{\sm$\mathrm{T_5}$[En]$+\mathrm{(TA)_5}$[EnCz]$+\mathrm{T_5}$[Cz]\normalsize}

\subsection{Extrinsic Evaluation: Machine Translation}
\label{sec:ex-eval}
In this section, we use multigraph embeddings to improve automatic translation between a source language $S$ and a target language $T$.
Multigraph model \mtmodel$\ $ uses:
text-based edges on a monolingual $S$ corpus ($\mathrm{T_5}[S]$),
text-and-alignment--based edges on a parallel corpus $S$-$T$ ($\mathrm{(TA)_5}[S$-$T]$);
and text-based edges on a monolingual $T$ corpus ($\mathrm{T_5}[T]$)~\footnote{
The two monolingual corpora should be from a fairly comparable domain.}.
The choice of the neighborhood creation is based on the results of the
previous evaluations, using a model optimized for multilingual semantic similarity
between $S$ and $T$.

\subsubsection{Inducing Translations from Non-parallel Data}
\label{sec:non-parallel-translations}
The empirical results we report here exploit the properties of our multilingual embeddings for
automatic translation purposes.
The intuition we intend to exploit is that words that are out-of-vocabulary
for the source-side of the parallel corpora (we denote them pOOV, for 'parallel Out-of-Vocabulary')
are embedded in the same multilingual embedding space as the words in the vocabulary of the target language
(both from the parallel-target and monolingual-target corpora).
Under a multilingual embedding space optimized for semantic similarity,
pOOV source words and their highest-cosine target-language neighbors tend to be
semantically related (for highest-cosine above a cosine threshold $T$);
therefore, such pairs should provide potentially correct translations, even for word-pairs
not directly observed in the parallel data.

To ground our intuition, we first present the results of a small evaluation done using WMT English-Spanish data.
We use a set of 100 word-pairs of the form $\langle w_s, w_t\rangle$,
randomly extracted according to the following set $C$ of criteria:
$w_s$ is an En pOOV in $\mathrm{WMT}$[EnEs], with a count of at least 100 in $\mathrm{Gigaword}$[En];
$w_t$ is the 1-highest (cosine-wise) Es neighbor of $w_s$ in \mtmodelenes,
with $\mathrm{cosine}(w_s, w_t)\ge T$ (we used $T=0.3$).
We ask human evaluators to rate each pair using one of the following ratings:
$\mathrm{similar}$ (well translated, with no/little semantic loss; syntactic differences are OK);
$\mathrm{hypernym}$ (translation is a hypernym (less specific); e.g.: "flower" is a hypernym for "daffodil");
$\mathrm{hyponym}$ (translation is a hyponym (more specific); e.g.: "catholic" is a hyponym for "churchgoer");
$\mathrm{other}$ (translation has other/no semantic relation with the source word).

We present in Table~\ref{tab:pOOV} the results of this evaluation.
The evaluation was done using two highly-skilled bilingual linguists, with (rare) disagreements resolved
in a second pass.
The results indicate that, in about 50\% of the cases, the translation proposed for the pOOV words are reasonable
(a hypernym is often a reasonable choice as a translation).

\begin{table}[th]
\begin{center}
\begin{tabular}{rrlr}
Rating              & \multicolumn{2}{c}{Example}       & Perc. \\ \hline
$\mathrm{similar}$  & \sm boozing$_{en}$   & \sm alcoholismo$_{es}$ & 15\% \\
$\mathrm{hypernym}$ & \sm lamprey$_{en}$   & \sm peces$_{es}$       & 35\% \\
$\mathrm{hyponym}$  & \sm preternatural$_{en}$ & \sm magia$_{es}$    &  4\% \\
$\mathrm{other}$    & \sm unladylike$_{en}$    & \sm arrogante$_{es}$& 46\% \\ \hline
\end{tabular}
\caption{Semantic-relation evaluation of English-Spanish pOOV (for $\mathrm{WMT}$[EnEs]) translations.}
\label{tab:pOOV}
\end{center}
\end{table}

\subsubsection{End-to-end Performance Impact using Translations from Non-parallel Data}
The results above are corroborated by quantitative evaluation results done using end-to-end MT systems.
To evaluate the impact of our technique in isolation, we use as base condition a non-neural,
phrase-based MT system.
We favor such a system in this evaluation simply because it allows a straightforward way
to evaluate how good the induced translations for the pOOV words are.

Under our base condition, we use phrase-based MT systems trained on
WMT data on three $S$-$T$ language pairs: English-Spanish, English-French, and English-Czech.
When possible, we use the publicly available monolingual and parallel data mentioned in Section~\ref{sec:data}:
the LDC Gigaword corpora for English, Spanish, and French; and
the WMT-2015 parallel corpora for EnEs, EnFr, and EnCz.
For monolingual Czech, we use a collection of 2.5 Bw of Czech news harvested from the web and automatically cleaned
(boiler-plate removed, formatting removed, encoding made consistent, etc.).
For each language, the target language model is trained on a monolingual corpus that subsumes
both the target-parallel and the target-monolingual corpora.

Under the test condition, we use the same system plus additional phrase-tables that
contain entries of the form $\langle w_s, w_t\rangle$ that meet the selection criteria $C$ from Section~\ref{sec:non-parallel-translations}.
For each language pair, we also create special test sets (see Table~\ref{tab:eval7}),
extracted from the English Gigaword to contain at least 1 pOOV word per sentence.
On these test sets, the decoder has the opportunity to use $w_t$ as a possible translation for $w_s$, in contrast to
the base condition in which the only choice of the decoder is to use $w_s$ in the target translation.

\begin{table*}[!th]
\begin{center}
\begin{tabular}{cccccc}
Model        & \multicolumn{2}{c}{pOOV Test}    & \multicolumn{3}{c}{Scores} \\
             & \sm Sentences & \sm Change-rate & \sm no-pOOV-transl & \sm pOOV-transl & \sm Delta [confidence] \\ \hline
\mtmodelenes & 1010          & 25.5\%          &  2.67              & 2.98            & +0.32 [0.19, 0.42] \\
\mtmodelenfr & 1001          & 62.2\%          &  2.81              & 2.96            & +0.14 [0.06, 0.22] \\
\mtmodelencz & 1004          & 20.7\%          &  1.94              & 2.19            & +0.25 [0.13, 0.36] \\ \hline
\end{tabular}
\begin{tabular}{rl}
\sm Example \#1 \\
\sm Source: & \fn The two sides have had a {\bf testy} relationship since Singapore was expelled from the Malaysian Federation.\\
\sm MT-Spanish Base:   & \fn  Las dos partes han mantenido una relaci\'{o}n {\bf testy} desde Singapur fue expulsada de la Federaci\'{o}n de Malasia. \\
\sm MT-Spanish Test:   & \fn  Las dos partes han tenido relaciones {\bf tensas} desde Singapur fue expulsada de la Federaci\'{o}n de Malasia. \\ \hline
\sm Example \#2 \\
\sm Source: &  \fn They doubted whether he could play the ultimate game, with the unique pressure it presented, with the \\
            &  \fn same {\bf elan} he displayed during the season as NFL Player of the Year. \\
\sm MT-Spanish Base:   & \fn  Ellos dudaron si podría jugar el partido final, con la única presión que presenta, con la misma {\bf Elan} se mostrará \\
            & \fn durante la temporada de la NFL como el Jugador del Año. \\
\sm MT-Spanish Test:   & \fn  Ellos dudaron si podría jugar el partido final, con la única presión que presenta, con la misma {\bf creatividad} que \\
            & \fn aparece durante la temporada de la NFL como el Jugador del Año. \\
\end{tabular}
\caption{Translation performance of multigraph bilingual models, using test sets from the English Gigaword
containing sentences with parallel-OOV (pOOV) terms; compares base condition (no pOOV translation)
against test condition (pOOV translations).
The evaluation uses a scale from 0:useless to 6:perfect. For instance, Example \#2 gets a score of 3 for the Base translation and a 5 for the Test-condition translation.}
\label{tab:eval7}
\end{center}
\end{table*}

As the results in Table~\ref{tab:eval7} show, the systems under the test condition score significantly higher compared to the systems under the base condition,
in a side-by-side, randomized blind comparison.
This evaluation is done using a pool of professional translators that see base-condition translations and test-condition translations side-by-side,
in a randomized position, without knowledge of what change in the MT system is being evaluated;
each evaluator is allowed to score only 1\% of the evaluation entries, to reduce bias;
the scale used is from 0 (useless) to 6 (perfect).
The values of the Delta numbers in Table~\ref{tab:eval7} are statistically significant at 95\%-confidence using bootstrap resampling~\cite{koehn:2004}.
The results indicate that the translations induced using \mtmodel$\ $ contribute significantly to the translation accuracy of sentences containing pOOV terms.
Moreover, the two examples in Table~\ref{tab:eval7} illustrate how our approach is orthogonal to other approaches proposed for dealing with rare or OOV words, such as
sub-word unit translations~\cite{sennrich-etal:16}.
Whereas for Example \#1 one can argue that a rare word like 'testy' could be decomposed as 'test'+'y' and subsequently translated correctly
using sub-word translations, this cannot happen with word 'elan' in Example \#2.
The proposed translation under the test condition, 'creatividad', may not be a perfect rendering of the word meaning for 'elan',
but it is superior to copying the word in the target translation (as under the base condition), or
to any potential sub-word--based translation.

\section{Conclusions and Future Work}
This article introduces a novel framework for building continuous word representations as $n$-dimensional real-valued vectors.
This framework utilizes multigraphs as an appropriate mathematical tool for inducing word representations from plain or annotated, monolingual or multilingual text.
It also helps with unifying and generalizing several previously-proposed word embedding models~\cite{mikolov-et-al:2013a,levy-goldberg:2014,luong-et-al:2015}.

We empirically show that this framework allows us to build models that yield word embeddings with
significantly higher accuracy on syntactic and semantic compositionality,
as well as multilingual semantic similarity.
We also show that the latter improvement leads to better translation lexicons for words that do not appear in the parallel training data.
The resulting translations are evaluated for end-to-end automatic translation accuracy,
and succeed in significantly improving the performance of an automatic translation system.

The family of models defined by our framework is much larger than what we presented here.
By choosing different neighborhood or objective functions, some of these models may further improve the performance on syntactic/semantic compositionality or similarity,
while others may have other properties yet to be discovered.
For instance, machine translation could potentially benefit from a thorough investigation study on the impact of various multigraph-based embeddings in pretraining,
which can incorporate a variety of signals (e.g., dependency information in source and/or target context).

In general, the high accuracy on the intrinsic tasks that some of these models exhibit, combined with their attractive computational costs,
makes them prime candidates for further exploring their properties and devising mechanisms to exploit them in end-to-end applications.


\end{document}